\documentclass[journal]{IEEEtran}

\usepackage{graphicx}
\usepackage{subfigure}
\usepackage{multirow}
\usepackage{multicol} 
\usepackage{threeparttable}
\usepackage{arydshln}
\usepackage{booktabs}
\usepackage[backref]{hyperref} 
\usepackage{CJK}
\usepackage{amsmath}
\usepackage{svg}
\usepackage{soul}
\usepackage{changepage}
\usepackage{listings}
\usepackage[ruled,vlined]{algorithm2e}

\begin{document}

\title{Negative Selection by Clustering for Contrastive Learning in Human Activity Recognition}

\author{Jinqiang Wang,
Tao Zhu,
Liming Chen,~\IEEEmembership{Senior Member,~IEEE,}
Huansheng Ning,~\IEEEmembership{Senior Member,~IEEE,}
and Yaping Wan
 
\thanks{Jinqiang Wang, Tao Zhu and Yaping Wan are with the School of Computer Science, University of South China, 421001 China. e-mail: tzhu@usc.edu.cn.}
\thanks{Liming Chen is with the Ulster University, Northern Ireland, UK. e-mail: l.chen@ulster.ac.uk}
\thanks{Huansheng Ning is with the School of Computer \& Communication Engineering, University of Science and Technology Beijing, 100083 China. e-mail: ninghuansheng@ustb.edu.cn}
}
\maketitle

\begin{abstract}
Contrastive learning has been applied to Human Activity Recognition (HAR) based on sensor data owing to its ability to achieve performance comparable to supervised learning with a large amount of unlabeled data and a small amount of labeled data. The pre-training task for contrastive learning is generally instance discrimination, which specifies that each instance belongs to a single class, but this will consider the same class of samples as negative examples. Such a pre-training task is not conducive to human activity recognition tasks, which are mainly classification tasks. To address this problem, we follow SimCLR to propose a new contrastive learning framework that negative selection by clustering in HAR, which is called ClusterCLHAR. Compared with SimCLR, it redefines the negative pairs in the contrastive loss function by using unsupervised clustering methods to generate soft labels that mask other samples of the same cluster to avoid regarding them as negative samples. We evaluate ClusterCLHAR on three benchmark datasets, USC-HAD, MotionSense, and UCI-HAR, using mean F1-score as the evaluation metric. The experiment results show that it outperforms all the state-of-the-art methods applied to HAR in self-supervised learning and semi-supervised learning.
\end{abstract}

\begin{IEEEkeywords}
Masking, Clustering, Negatives, Contrastive Learning, Human Activity Recognition, Sensor Data.
\end{IEEEkeywords}

%
\IEEEpeerreviewmaketitle

\section{Introduction}
%
%
%
%

\IEEEPARstart{T}{he} development of human activity recognition (HAR) technology based on wearable sensors (accelerometers, gyroscopes) has contributed to advances in the fields of smart homes \cite{rashidi2009keeping}, fall detection \cite{tsinganos2018comparison} \cite{ngu2017fall} and healthcare rehabilitation \cite{patel2012review} \cite{zhou2020deep}. In recent years, deep learning techniques \cite{zhang2020sensors} \cite{murahari2018attention} \cite{zhang2019novel} \cite{zhao2019indoor} applied to human activity recognition have substantially improved activity recognition accuracy compared to traditional machine learning \cite{lara2012survey} \cite{chathuramali2012faster} \cite{ramasamy2018recent}. However, supervised learning usually requires a large number of labeled data sets to train the activity recognition model and generally requires manual labeling of sensor data. This process is time-consuming and tedious, especially in healthcare, where labeled data is more challenging to collect. Moreover, the labels are affected by various noise sources, such as sensor noise, segmentation problems, and changes in the activities of different people, making the annotation process error-prone \cite{chen2021deep}. Therefore, insufficient data annotation becomes a major challenge for HAR.
\par
To alleviate the problem of insufficient data annotation, contrastive learning, a paradigm of self-supervised learning, has achieved excellent performance in computer vision \cite{le2020contrastive}. The pre-training process for contrastive learning is to generate pseudo-labels using data augmentation on a large amount of unlabeled data, enabling the model to learn to distinguish which augmented versions are positive pairs and which are negative pairs \cite{bachman2019learning}. The pre-trained learned model is fine-tuned in downstream tasks using a small amount of labeled data to achieve performance comparable to supervised learning \cite{jaiswal2020survey} \cite{falcon2020framework}. 
There are many types of pre-training tasks for contrastive learning, such as MoCo \cite{he2020momentum} \cite{chen2020improved} and SimCLR \cite{chen2020simple} \cite{chen2020big} with instance discrimination \cite{wu2018unsupervised} as the task, and NNCLR \cite{dwibedi2021little}, MSF \cite{koohpayegani2021mean}, TTL \cite{wang2021solving} and HardCL \cite{robinson2020contrastive} which redefine positive and negative pairs based on the instance discrimination task. In addition to this, SwAV \cite{caron2020unsupervised} uses clustering to reduce feature dimensionality, BYOL \cite{grill2020bootstrap} and SimSiam \cite{chen2021exploring} that drop the use of negative examples and use similarity metrics for the pre-training task.
In summary, the pre-training task of the above work is to enable the model to generate representations that are similar for positive example pairs and distant for negative example pairs in the latent space. However, the pre-training task of these works is essentially individual discrimination, which requires that augmented samples that do not belong to the same instance all constitute negative pairs. It is likely to pull apart sample representations belonging to the same class but not the same instance.
\par
\begin{figure*}[htbp]
\centering
\subfigure[over clustering]{
\label{fig:subfig:a}
\begin{minipage}[t]{0.5\linewidth}
  \centering
  \includegraphics[width=2in]{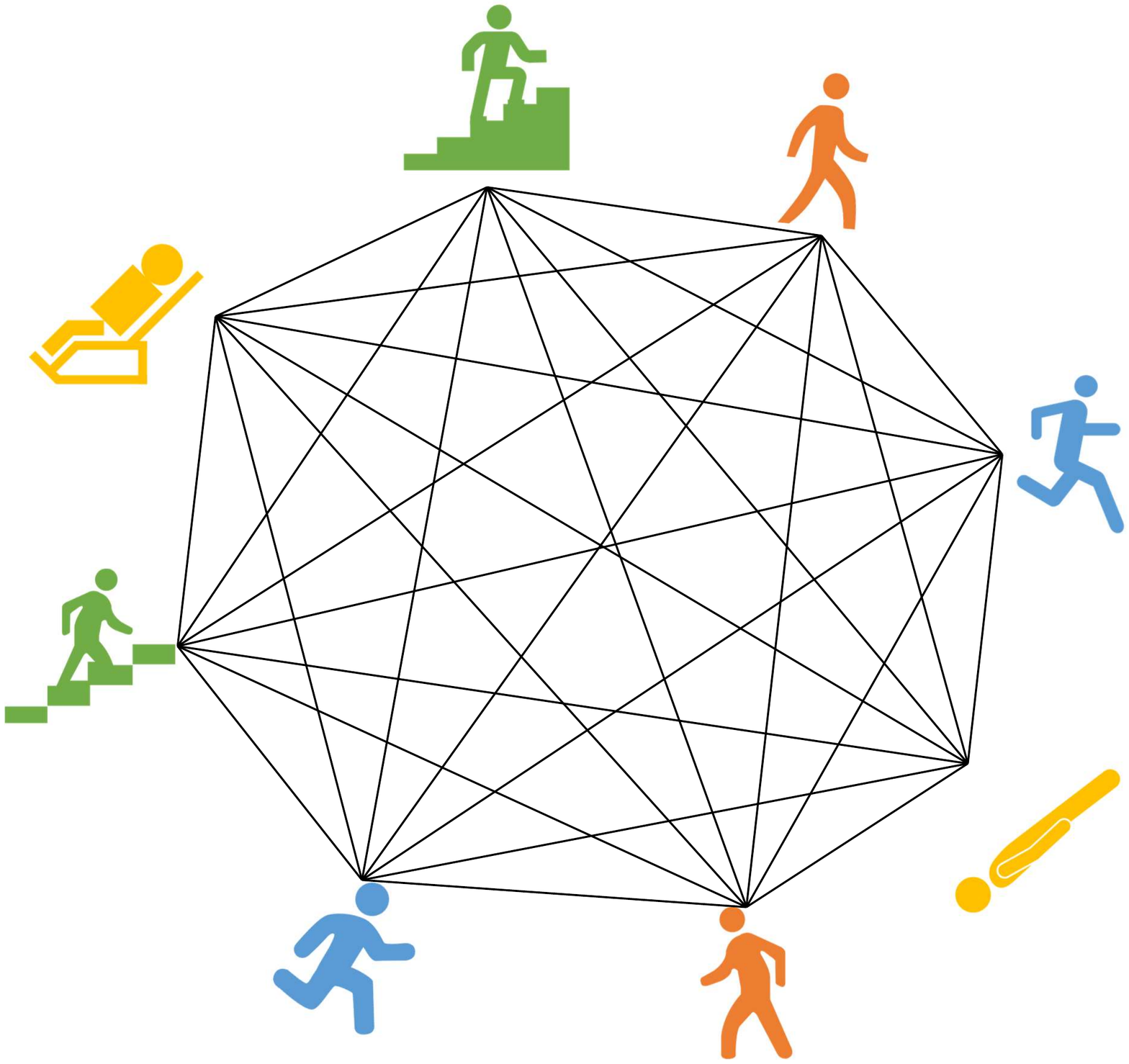}
\end{minipage}%
}%
\subfigure[ideal result]{
\label{fig:subfig:b}
\begin{minipage}[t]{0.5\linewidth}
  \centering
  \includegraphics[width=2in]{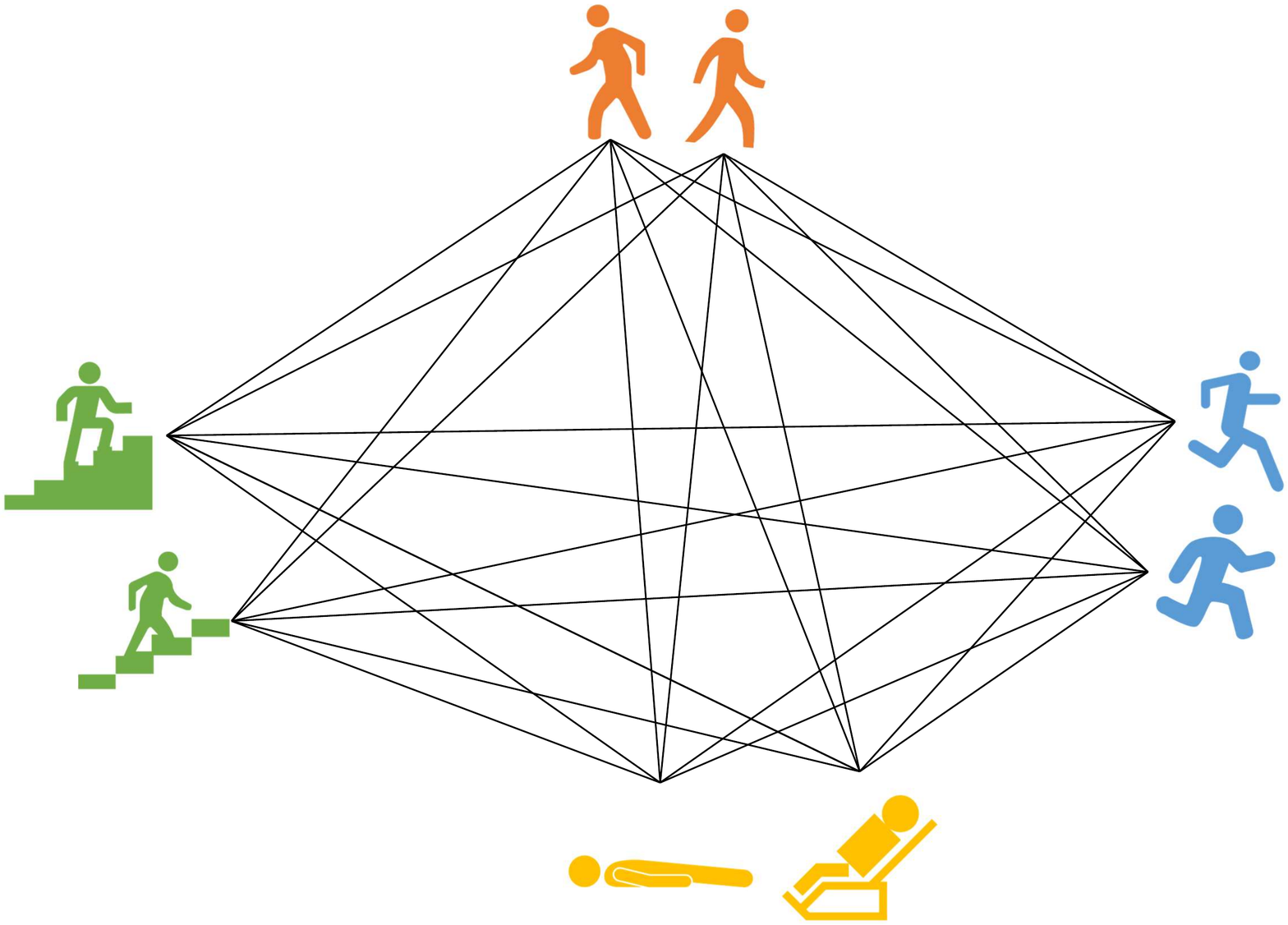}
\end{minipage}
}%
\caption{Activity Representations in Latent Space}
\label{fig:1}
\end{figure*}
The pre-training tasks currently applied to contrastive learning in human activity recognition are also instance discrimination \cite{wu2018unsupervised}, such as SimCLRHAR \cite{tang2020exploring} and CSSHAR \cite{khaertdinov2021contrastive} based on SimCLR \cite{chen2020simple} improvement and MoCoHAR \cite{wang2021sensor} based on MoCo \cite{chen2020improved} improvement. These works generally optimize on augmentation methods and backbone networks and do not improve on pre-training tasks. Contrastive learning models that use instance discrimination as a pre-training task tend to fall into over clustering \cite{wang2021solving} during training. As shown in Fig. \ref{fig:subfig:a}, each activity is classified into a single class in the latent space, due to the fact that the instance discrimination task specifying that representations that do not belong to the same instance will be pulled apart. Such representations are not conducive to the downstream classification task. As shown in Fig. \ref{fig:subfig:b}, our ideal result is that the same class of activities is represented similarly in the latent space. Most of the human activity recognition tasks based on sensor data are classification tasks, and contrastive learning with instance discrimination as a pre-training task contradicts the classification tasks. Therefore, it is a challenge to avoid distancing the same class of sample representation when calculating the contrastive loss (instance discrimination). In other words, how to avoid treating the same class of samples as negative examples based on the instance discrimination task is the current problem to be solved.
\par
\begin{figure*}[htbp]
\centering
\subfigure[SimCLR]{
\label{fig:subfig:c}
\begin{minipage}[t]{0.5\linewidth}
  \centering
  \includegraphics[width=0.8\linewidth]{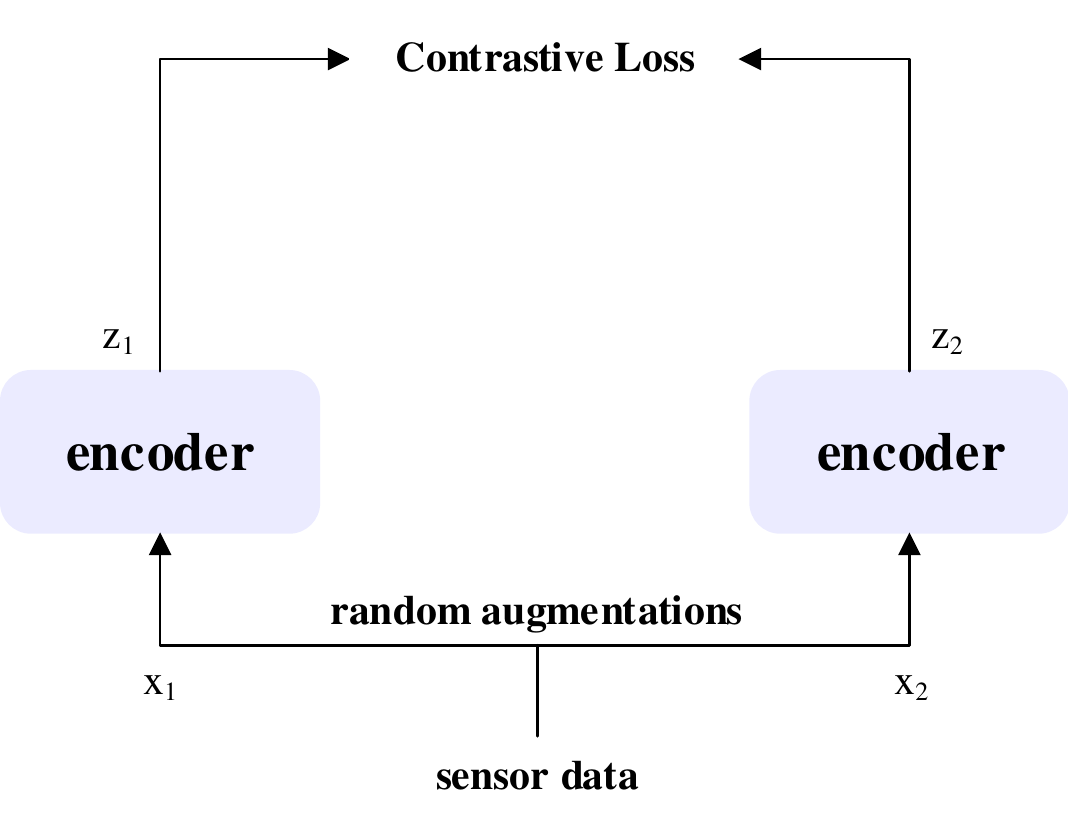}
\end{minipage}%
}%
\subfigure[ClusterCLHAR]{
\label{fig:subfig:d}
\begin{minipage}[t]{0.5\linewidth}
  \centering
  \includegraphics[width=0.8\linewidth]{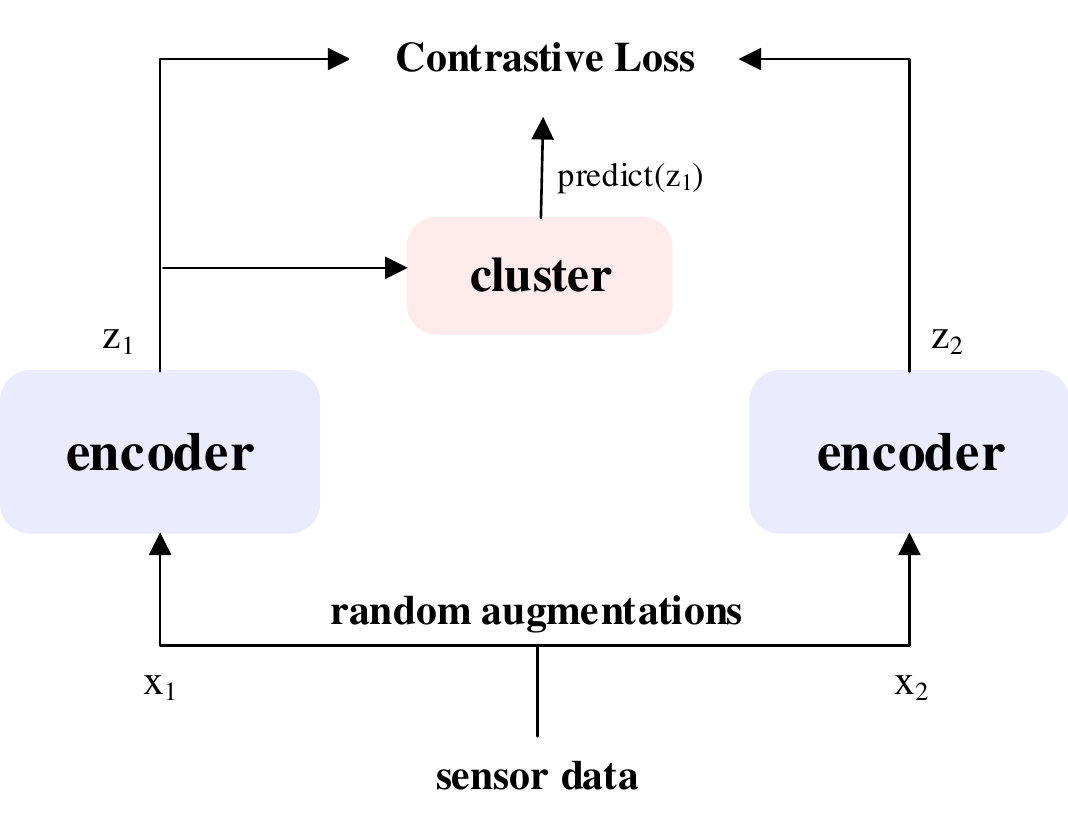}
\end{minipage}
}%
\caption{The Outline of Contrastive Learning Frameworks }
\label{fig:2}
\end{figure*}
To address the problem of considering the same class of samples as negative examples, we follow SimCLR \cite{chen2020simple} proposed a new contrastive learning framework that negative selection by clustering in HAR, ClusterCLHAR, as shown in Fig. \ref{fig:subfig:d}. Compared with SimCLR (Fig. \ref{fig:subfig:c}), we designed a new contrastive loss function (pre-training task). It uses unsupervised clustering techniques based on the instance segmentation task to mask the same-cluster samples to the extent that the same-cluster samples obtained from clustering are not considered negative examples. Namely, the same-cluster samples obtained from clustering can be ignored in the calculation of cross-entropy. This will minimize the situation of considering the same class of samples as negative examples.

\par
To evaluate the performance of the ClusterCLHAR framework, we use TPN \cite{saeed2019multi} as the backbone network, which uses a convolution-based network structure, is fast in inference, and outperforms DeepConvLSTM \cite{ordonez2016deep} in activity recognition performance.
Three benchmark datasets, USC-HAD \cite{zhang2012usc}, MotionSense \cite{malekzadeh2018protecting}, and UCI-HAR \cite{anguita2013public}, were used for the experiment. The experiments are first compared to previous work on self-supervised learning (downstream tasks using a large amount of labeled data), and the results show that ClusterCLHAR outperforms all state-of-the-art self-supervised learning work on three datasets. In addition, we evaluate ClusterCLHAR on semi-supervised learning (downstream tasks using a small amount of labeled data). This part of the experiment focuses on comparing supervised learning and previous contrastive learning methods. The pre-trained model was fine-tuned using 1\% and 10\% of the labeled data in the downstream task, respectively, and the remaining data were used as the test set. The experiment results show that ClusterCLHAR outperforms all state-of-the-art contrastive learning methods on three datasets.
\par
The contributions of this paper are as follows: 1. A new contrastive learning framework ClusterCLHAR is proposed, which uses a clustering method to filter the negative samples compared to SimCLR and is more compatible with the classification task. 2. The impact of the confidence level of the clustering results on the model performance is analyzed in detail. 3. The details of the model (clustering method, number of clusters) are further discussed.
\par
The remainder of this paper is structured as follows. In Section II, the discussion of the work on contrastive learning negative example selection and the work on contrastive learning applied to HAR is presented. In Section III, the ClusterCLHAR framework is described in detail. In Section IV, self-supervised learning and semi-supervised learning experiment protocols are designed to evaluate the performance of our proposed framework. In Section V, the main experiment results are presented and discussed. In Section VI, the framework details are further studied. In Section VII, the paper is summarized and future work is proposed based on the identified shortcomings.

\section{Related Works}

\subsection{Definition of contrastive learning negative examples}
Contrastive learning, a paradigm of self-supervised learning, uses data augmentation to generate pseudo-labels that enable the model to distinguish between positive and negative pairs in augmented samples \cite{bachman2019learning}. Contrast learning has three steps \cite{jaiswal2020survey}, the first step is the augmentation method, which determines the quality of pseudo-label generation and has a significant impact on the final performance of the model. The second step is the encoder, which encodes the augmented samples and affects the quality of the generated representation in the latent space. The third step is the loss function (pre-training task), which defines the positive and negative samples that determine in which direction the model will learn. 
\par
The definition of negative examples plays a crucial role in the final performance of contrastive learning, and many contrastive learning works have been improved on this issue. MoCo \cite{he2020momentum} uses queues to expand negative examples and achieve excellent performance with a small batch size. SimCLR \cite{chen2020simple} uses a larger batch size to expand the number of negative examples, which improves the performance of contrastive learning in a simple and efficient way. NNCLR \cite{dwibedi2021little} compares with SimCLR to select the most similar sample representation from a queue by the nearest neighbor method instead of the original representation to calculate the contrastive loss, thus improving the model performance by increasing the training complexity. MSF \cite{koohpayegani2021mean} compared to SimCLR calculates the contrastive loss by selecting the K most similar sample representations in a queue and calculating the mean instead of the original sample representation. TTL \cite{wang2021solving} and HardCL \cite{robinson2020contrastive} define near positive examples as negative examples for the purpose of separating similar sample representations in the latent space.
\par
However, with the downstream task identified as a classification task, the above work is contradictory for how negative examples are defined because this definition would consider each instance as a single class. We expect that the same class of sample representations remains close while the different classes are pushed apart in the latent space. For this reason, when designing the loss function, the possibility of pushing them apart in the latent space is reduced if we can try to avoid considering same class samples as negative examples. The resulting encoder representation can approach the effect of Fig. \ref{fig:subfig:b}, which will be easier and more efficient for training of downstream classification tasks.

\subsection{Contrastive learning for HAR}
Some self-supervised learning works are applied to human activity recognition based on sensor data, such as Multi-task SSL \cite{saeed2019multi}, CAE \cite{haresamudram2019role}, Masked Reconstruction \cite{haresamudram2020masked}, and CPCHAR \cite{haresamudram2021contrastive}. These works use the data to generate pseudo labels and set pretext tasks based on the pseudo labels, enabling the encoder to learn excellent representations by completing the pretext tasks. Contrastive learning as a paradigm of self-supervised learning has been applied to human activity recognition. SimCLRHAR \cite{tang2020exploring} first applies contrastive learning to HAR, using SimCLR's architecture, and achieves a slight improvement in recognition accuracy relative to supervised learning activities. CSSHAR \cite{khaertdinov2021contrastive} replaces the backbone network with a custom Transformer. Although the accuracy is improved overall, the difference in performance relative to the supervised learning of the backbone network is not significant. MoCoHAR \cite{wang2021sensor} takes the sensor data augmentation method of contrastive learning as an entry point and proposes resampling data augmentation, which improves the accuracy significantly with a small amount of labeled data compared to supervised learning. However, the contrasting loss functions of the above work are InfoNCE \cite{van2018representation} and NT-Xent \cite{sohn2016improved}, which are tasked with instance discriminations and will consider the same class of samples as negative examples. In particular, human activity recognition tasks are mostly classification tasks, but pre-training tasks such as the instance discrimination cause the representation of the same class of activities to be pulled apart in the latent space. 
To address this problem, we improve the contrastive loss function by introducing an unsupervised clustering technique that discriminates and masks the same cluster samples. This approach avoids the same cluster of samples representation as negative examples to the extent that the same cluster of samples in negative examples is ignored in the calculation of cross-entropy losses. This will possibly make the representation of the same class of activities closer in the latent space.

\section{Methods}
\subsection{Framework}

When studying the contrastive loss functions InfoNCE \cite{van2018representation} and NT-Xent \cite{sohn2016improved} with individual discrimination as the task, it was found that they consider the same class samples as negative examples. The resulting encoder representation is likely to be assigned to a single class for each instance in the latent space, which conflicts with the representation needed in the downstream classification task. In order to make the pre-training task objective as similar as possible to the downstream classification task, we mask the same class of samples in the negative example samples based on NT-Xent to avoid the same class of samples being considered as negative example samples. In an unsupervised learning environment, unsupervised clustering methods are used to mask sample representations that may be of the same cluster in negative example samples. Here we use clusters to approximate the true classification. We cluster a branch's sample representations to obtain the same cluster markers, and the same cluster will no longer be considered negative examples when performing contrastive losses. Thus we propose a new contrastive learning framework based on human activity recognition called ClusterCLHAR.

\begin{figure*}
  \centering
  \includegraphics[width=1\linewidth]{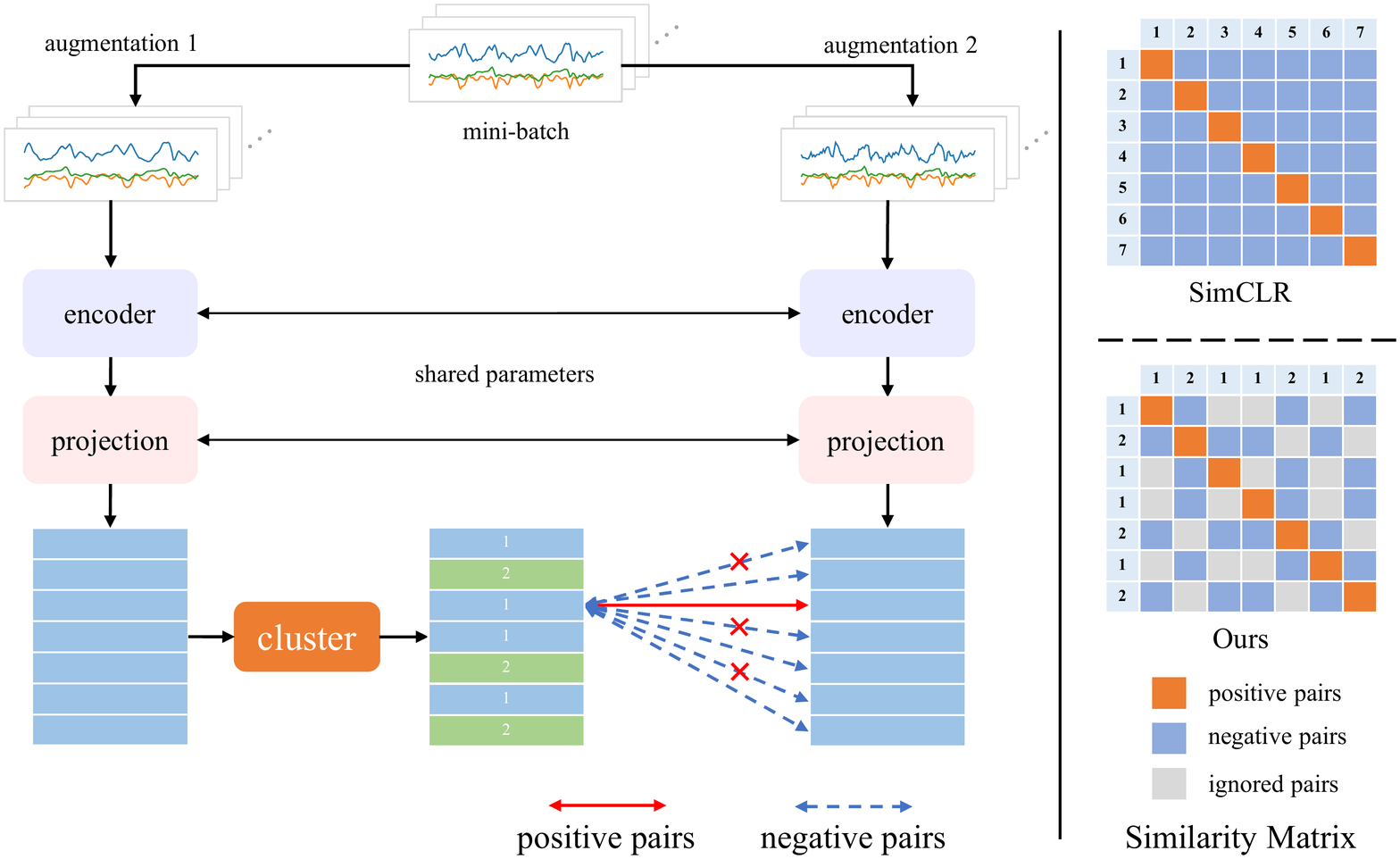}
\caption{Negative Selection by clustering for Contrastive
Learning in Human Activity Recognition (ClusterCLHAR)}
  \label{fig:3}
\end{figure*}
\par 
The flow chart of the framework is shown in Fig. \ref{fig:3}. For a mini-batch of data, firstly, two different sets of samples are generated by different data augmentation, and then two different sets of sample representations are obtained by encoder and projection header.
\par
For the sample representation of the two branches, SimCLR uses NT-Xent as the contrasting loss function. The implementation is shown in Eq. \eqref{eq1}, where $(i,j)$ is a positive pair, $z_i$ denotes the representation output by the projection head, $N$ denotes the mini-batch length, $I_{[\cdot]}$ is the judgment function, which is equal to 1 when the expression in $[\cdot]$ is true and 0 vice versa, and $\tau$ is the temperature coefficient. From the formula, it can be seen that sample i constitutes a negative pair for all samples except sample j. This means that sample i will regard samples of the same class as negative examples, which is incompatible with the classification task.
\begin{equation}
   l_{i,j}^{NT-Xent}=-\log {\frac{exp(z_i \cdot z_j / \tau )}{\sum_{k=1}^{2N}I_{[k\neq i]}exp(z_i \cdot z_k /\tau)}} 
   \label{eq1}
\end{equation}
\par
To address this problem, in the context of unsupervised learning, we introduce clustering methods that will be used in a branch-generated representation. As shown in Fig. \ref{fig:3}, we have clustered the representations generated by the first branch so that each representation is assigned a cluster label. The definition of positive pairs is the same as NT-Xent. However, the difference is that the sample representation of the same cluster is ignored in the definition of negative pairs.
A new contrast loss function Cluster-NT-Xent is proposed to systematically describe our method, as shown in Eq. \eqref{eq2}, where cluster(i) denotes the set of clusters in which sample i is located. Unlike Eq. \eqref{eq1}, it does not constitute negative sample pairs for sample representations assigned to the same cluster, as shown in the right part of Fig. \ref{fig:3}. The batch calculation is shown in Eq. \eqref{eq3}.
\begin{equation}
   l_{i,j}^{Cluster}=-\log {\frac{exp(z_i \cdot z_j / \tau )}{\sum_{k=1}^{2N}I_{[k\notin cluster(i)]}exp(z_i \cdot z_k /\tau)}} 
   \label{eq2}
\end{equation}
\begin{equation}
   \mathcal{L}^{Cluster}=\frac{1}{2N} \sum_{k=1}^{N}[l_{2k-1,2k}^{Cluster}+l_{2k,2k-1}^{Cluster}]
   \label{eq3}
\end{equation}
The overall MaskCLRedHAR process is as follows.
\par \textbf{Data Augmentation:} Using resampling \cite{wang2021sensor} as the augmentation method of this framework, one branch does not use data augmentation and uses the original samples directly, while the other branch uses data augmentation.
\par \textbf{Encoder:} A TPN \cite{saeed2019multi} is used as the encoder, which is an inference-quick framework for human activity recognition. The encoder encodes two sets of samples generated by different data augmentations, and the two branch encoder parameters are shared.
\par \textbf{Projection Head:} Based on the experience of work \cite{chen2020simple}, we use a nonlinear projection head to remap the representation generated by the encoder to a new dimensional representation. The projection head parameters are shared. 
\par \textbf{Contrastive Loss Function:} Eq. \eqref{eq3} is used as the contrastive loss function of this framework.
\par The ClusterCLHAR pre-training pseudo-code is shown in Algorithm \ref{alg:code}.

\begin{algorithm}[t]
\caption{ClusterCLHAR Pseudocode.}
\label{alg:code}
\definecolor{codeblue}{rgb}{0.25,0.5,0.5}
\lstset{
  backgroundcolor=\color{white},
  basicstyle=\fontsize{7.2pt}{7.2pt}\ttfamily\selectfont,
  columns=fullflexible,
  breaklines=true,
  captionpos=b,
  commentstyle=\fontsize{7.2pt}{7.2pt}\color{codeblue},
  keywordstyle=\fontsize{7.2pt}{7.2pt},
}
\begin{lstlisting}[language=python]
# f, g: encoder and projection head
# N: batch size
# t: temperature
# cluster: clustering methods, such as K-means

LARGE_NUM=1e9
for x in loader:  # load a minibatch x with N samples
    x1, x2 = aug(x), aug(x)    # random augmentation

    z1, z2 = g(f(x1)), g(f(x2))  #forward   
    
    p1 = normalize(z1)
    p2 = normalize(z2)
    
    labels = range(N)
    loss_a = CrossEntropyLoss(sim(p1,p2), labels)
    loss_b = CrossEntropyLoss(sim(p2,p1), labels)
    loss = loss_a + loss_b
    
    loss.backward() # back-propagate
    update([f.params, g.params]) # Adam update
        
# similarity of positive and negative  pairs
def sim(p1,p2): 
    logits_ab = matmul(p1, p2.T)/t  #(N,N)
    logits_aa = matmul(p1, p1.T)/t  #(N,N)
    
    # mask generation by clustering
    p1_clu = cluster.fit_predict(p1.cpu()) #(N,)
    
    masks_aa = [i == p1_clu for i in p1_clu] #(N,N)
    logits_aa = logits_aa - masks_aa * LARGE_NUM  #(N,N)
    
    masks_ab = masks_aa - eye(N)   #(N,N)
    logits_ab = logits_ab - masks_ab * LARGE_NUM  #(N,N)
    
    return concat([logits_ab, logits_aa], axis=1)
\end{lstlisting}
\end{algorithm}

\subsection{Clustering Confidence}\label{Clustering Confidence}
The performance of clustering methods has a significant impact on model performance. If the clustering results differ significantly from the true classification, then many true negative pairs will be ignored, seriously affecting the pre-training performance. When using unsupervised clustering methods, the samples at the boundary of each cluster are usually considered as negative examples or anomalous samples \cite{rokach2005clustering}. We decided to introduce this concept in our proposed approach to verify whether it affects the performance of the framework. By setting a certain percentage of samples closest to each cluster center as normal samples and the rest as anomalous samples, we will set the anomalous samples as a single cluster.
\begin{equation}
   l_{i,j}^{confidence}=-\log {\frac{exp(z_i \cdot z_j / \tau )}{\sum_{k=1}^{2N}I_{[k\notin cluster(i)\cup \neg thr(\alpha,i)]}exp(z_i \cdot z_k /\tau)}} 
   \label{eq4}
\end{equation}
\par The implementation is shown in Eq. \eqref{eq4}, where thr($\alpha$,i) indicates that the value is 1 if sample i belongs to the top $\alpha$\% of the sample representation closest to the cluster center, and 0 otherwise. Here $\alpha \in [0,100]$,  denotes the confidence level on the clustering results. When $\alpha$ decreases, the number of samples in each cluster decreases and the number of negative pairs.  increases. In simple terms, the set of sample representations in the top $\alpha$\% of the cluster center will calculate the loss according to Eq. \eqref{eq2}, while representations that do not belong to this set will calculate the loss according to NT-Xent. When $\alpha$ equals 100, the whole mathematical model degenerates to Eq. \eqref{eq2}, and when $\alpha$ equals 0, the whole equation degenerates to NT-Xent.

\section{Experiment}
\subsection{Datasets}
\par
The USC-HAD \cite{zhang2012usc} dataset was collected on the MotionNode sensing platform and contained accelerometer and gyroscope data. This dataset consists of data from 14 subjects recording 12 activities, including walking forward, walking left, walking right, going upstairs, going downstairs, running forward, jumping, sitting, standing, sleeping, and riding the elevator up and down.  All data were collected at a 100 Hz sampling rate.
\par
The MotionSense \cite{malekzadeh2018protecting} dataset consists of time-series data generated by accelerometer and gyroscope sensors. An iPhone 6s was placed in the participant's front pocket and information was collected from the core motion framework on the iOS device using SensingKit. All data was collected at a 50 Hz sampling rate. A total of 24 participants of different genders, ages, weights, and heights performed six activities: downstairs, upstairs, walking, jogging, sitting, and standing in 15 trials under the same environment and conditions.
\par
The UCI HAR \cite{anguita2013public} activity recognition dataset was collected from 30 subjects who performed basic activities and postural transitions while carrying a waist-mounted smartphone with embedded inertial sensors. Six basic activities were included: standing, sitting, lying, walking, upstairs and downstairs. Experiments captured 3-axis linear acceleration and 3-axis angular velocity at a constant 50 Hz rate using the device's built-in accelerometer and gyroscope.
\par
The experiments in this paper will use the accelerometer and gyroscope data from the above datasets.

\subsection{Self-supervised experiment protocol}
According to work \cite{saeed2019multi} \cite{tang2020exploring}, the USCHAD and MotionSense datasets were segmented with 400 sample points as a sliding window with 50\% overlap between windows. According to work \cite{anguita2013public} \cite{wang2021sensor}, the UCI-HAR is segmented with 128 sample points as a sliding window with 50\% overlap between windows. Based on experimental protocols from previous work \cite{haresamudram2019role} \cite{saeed2019multi}, USC-HAD sensor data from subjects 11 and 12 was used for validation, while data from subjects 13 and 14 was used as a test set. The splitting protocol of  MotionSense and UCI-HAR are 20\% of subjects are randomly sampled for the test split, while 20\% of the remaining subjects are selected for the validation set.
\par
In this paper, the contrastive learning pre-training task takes TPN as the backbone and adds three projection heads with dimensions of 96,96, and 96, respectively. The optimizer uses Adam \cite{kingma2014adam} with an initial learning rate of 1e-3. The temperature coefficient is 0.1, the batch size is 1024, and the model is trained for 200 epochs. All deep learning code is built on the TensorFlow \cite{abadi2016tensorflow} platform. An NVIDIA GeForce RTX 3090 GPU was used to accelerate the training process.
The loss function is based on Eq. \eqref{eq3}. The clustering method used in USC-HAD and MotionSense is K-means \cite{macqueen1967some} \cite{sculley2010web}, and the clustering method used on UCI-HAR is BIRCH \cite{zhang1996birch}. The number of clustering centers defaults to the true number of classifications in the dataset. The pre-training task uses a single dataset divided into parts of the training set, and the network parameters are initialized using randomization.
When pre-training performance is evaluated in the downstream activity recognition task, the pre-trained model is thrown off the projection header, keeping only the encoder and freezing all layers, with a trainable linear classification layer added at the end of the model. The optimizer uses Adam with an initial learning rate of 10. The model is trained for 200 epochs, using the mean F1-score \cite{powers2020evaluation} as the evaluation metric. All experiments were trained ten times, and the results were averaged.

\subsection{Semi-supervised experiment protocol}
This part of the experiment is designed to simulate the performance of our proposed framework in the context of insufficient labels. In the data preprocessing stage, we narrowed the sliding window to be more relevant to the situation of insufficient labels. Based on previous work \cite{wang2021sensor}, the USC-HAD and MotionSense datasets were segmented using a sliding window of 200 samples, with 25\% overlap for USCHAD and 12.5\% overlap for MotionSense. The UCI-HAR dataset maintains the settings of the self-supervised experiment protocol. To simulate the situation of insufficient labels, this paper uses two training set proportion settings of 1\% and 10\% randomly selected in the downstream task.
\par
The pre-training phase uses all unlabeled data from a single dataset, and the rest is identical to the self-supervised experiment protocol. To evaluate the performance of contrastive learning pre-training, two evaluation protocols are used in the downstream task. 
\par Linear evaluation: Freeze all layers of the encoder and add a trainable linear classification layer at the end of the model. The optimizer uses Adam with an initial learning rate of 1e-1.
\par Fine-tuning: Unfreeze the last two layers of the encoder and add a trainable linear classification layer to the end of the model. The optimizer uses Adam with an initial learning rate of 1e-2.
\par
The loss function uses cross-entropy. The batch size is 50 and 500 according to the proportion of 1\% and 10\% of different training sets, respectively. The model is trained for 200 epochs, and the mean F1-score is used as the evaluation metric. All experiments were trained ten times, and the results were averaged.

\section{Results}
\subsection{Self-supervised learning}
In this section of experiments, we evaluate the performance of ClusterCLHAR in the activity classification task. We compare with state-of-the-art self-supervised learning work, where the training and test sets are divided in the same way. The performance of the TPN backbone network under supervised learning was also compared. The experimental results are shown in Table \ref{Tab01}, and the results of the state-of-the-art work are from \cite{haresamudram2020masked} \cite{khaertdinov2021contrastive}.
\begin{table}
  \centering
  \caption{Self-supervised learning}
  \label{Tab01}
  \setlength{\tabcolsep}{1.5mm}{
  \begin{tabular}[]{lcccc}
    \toprule
      Method &Type &USC-HAD &MotionSense &UCI-HAR\\
    \midrule
    TPN \cite{saeed2019multi} &Sup. &55.60 &93.00 &94.27\\
    \midrule
    Multi-task SSL \cite{saeed2019multi} &SSL &45.37 &83.30 &80.2\\
    CAE \cite{haresamudram2019role} &SSL &48.82 &82.50 &80.26\\
    Masked Reconstruction \cite{haresamudram2020masked} &SSL &49.31 &88.02 &81.89\\
    CPCHAR \cite{haresamudram2021contrastive} &SSL &52.01 &- &81.65\\
    CSSHAR \cite{khaertdinov2021contrastive} &SSL &57.76 &- &91.14\\
    ClusterCLHAR (ours) &SSL &\textbf{58.85} &\textbf{89.22} &\textbf{92.12}\\
    \bottomrule
  \end{tabular}
  }
\end{table}
\par
The experiment results show that our proposed ClusterCLHAR outperforms all state-of-the-art self-supervised learning methods on all three benchmark datasets. This can demonstrate that our proposed framework can better capture data representations on unlabeled data and be applied to downstream activity recognition tasks. Notably, ClusterCLHAR outperforms supervised learning of the corresponding backbone network on the USCHAD dataset, which no other method can compete with. ClusterCLHAR also closed the gap with supervised learning on the other two datasets, which validates our hypothesis that the contrastive learning pretext task is compatible with the downstream activity recognition task.

\subsection{Semi-supervised learning}
In this subsection, we evaluate the performance of our proposed method in the context of contrastive learning and a small amount of labeled data. We use TPN as a benchmark for supervised learning and compare two contrastive learning works, SimCLRHAR \cite{tang2020exploring} and MoCoHAR \cite{wang2021sensor}. For a fair comparison, we use resampling data augmentation for SimCLRHAR.
In addition, we extend NNCLR \cite{dwibedi2021little} with the same data augmentation method, encoder, and projection head as our framework and add it to the comparison. Note that our comparison focuses on the difference of model performance by the negative example definition method in contrastive learning. The experiment results are shown in Table \ref{Tab02}.
\begin{table*}[]
  \centering
  \caption{Semi-supervised learning}
  \label{Tab02}
  \setlength{\tabcolsep}{2mm}{
  \begin{tabular}[]{lcccccccccccc}
    \toprule
    \multirow{3}{*}{} 
		&\multicolumn{4}{c}{USC-HAD} &\multicolumn{4}{c}{MotionSense} &\multicolumn{4}{c}{UCI-HAR}\\
		\cmidrule(r){2-5}\cmidrule(r){6-9}\cmidrule(r){10-13}
		& \multicolumn{2}{c}{1\%} & \multicolumn{2}{c}{10\%} 
		& \multicolumn{2}{c}{1\%} & \multicolumn{2}{c}{10\%}
		& \multicolumn{2}{c}{1\%} & \multicolumn{2}{c}{10\%}\\
		
		\cmidrule(r){2-3} \cmidrule(r){4-5}\cmidrule(r){6-7} \cmidrule(r){8-9}
		\cmidrule(r){10-11} \cmidrule(r){12-13}
		
		&Linear. &Fine.  &Linear. &Fine. &Linear. &Fine.  &Linear. &Fine.
		&Linear. &Fine.  &Linear. &Fine.\\
    \midrule
    TPN (Sup.) \cite{saeed2019multi} &\multicolumn{2}{c}{70.23} &\multicolumn{2}{c}{85.93} &\multicolumn{2}{c}{84.91} &\multicolumn{2}{c}{95.06} &\multicolumn{2}{c}{90.50} &\multicolumn{2}{c}{95.47} \\
    SimCLRHAR \cite{tang2020exploring} &44.84 &63.82 &53.16 &83.32 &79.16 &84.62 &78.26 &95.77 &54.52 &62.44 &59.14 &78.27\\
    SimCLRHAR(resampling \cite{wang2021sensor}) &71.48 &72.87 &83.51 &85.50 &83.76 &86.55 &92.60 &96.08 &83.53 &86.75 &92.91 &95.58 \\
    MoCoHAR \cite{wang2021sensor} &68.60 &69.20 &78.51 &85.84 &77.66 &85.49 &91.72 &96.63 &83.56 &87.41 &91.89 &95.49\\
    NNCLR \cite{dwibedi2021little} &72.36 &74.70 &83.81 &87.07 &83.39 &86.19 &92.09 &96.40 &83.89 &87.53 &95.59 &95.55 \\
    ClusterCLHAR (ours) &\textbf{76.08} &\textbf{78.09} &\textbf{85.01} &\textbf{87.86} &\textbf{86.69} &\textbf{87.35} &\textbf{94.02} &\textbf{96.44} &\textbf{88.78} &\textbf{90.53} &\textbf{94.68} &\textbf{95.91} \\
   \bottomrule
  \end{tabular}
  }
\end{table*}
\par
The experiment results show that our proposed framework outperforms all state-of-the-art contrastive learning methods in both linear evaluation and fine-tuning on all three datasets. With 1\% labeled data, ClusterCLHAR  outperforms the previous best contrastive learning work by 3.72\% on USC-HAD and 3.3\% on Motion Sensor and 4.89\% on UCI-HAR with linear evaluate. It outperforms the previous best work by 3.39\% on USC-HAD and 1.16\% on Motion Sensor and 3\% on UCI-HAR with fine-tune. SimCLR(resampling) is most similar to our work, the only difference is that the contrastive loss function (pre-training task).  Our method outperformed SimCLR (resampling) by 4.6\%, 2.93\%, and 5.25\%, respectively, on the three datasets under linear evaluation of 1\% labeled data. This demonstrates that our approach of masking same-cluster negative examples used in the contrasting loss function has a positive impact on the downstream classification task. Notably, with 1\% labeled data, our method performs better than supervised learning in both linear and fine-tuned evaluations on the USCHAD and MotionSense datasets. However, our approach achieves competitive performance with supervised learning on UCI-HAR.  We analyze that this is because, under supervised learning, the model is easier to train in the UCI-HAR dataset relative to other datasets. This result compresses the improvement space of comparative learning, making contrastive learning models without an explicit task insignificantly worse than supervised learning with an explicit classification task.

\section{Discussion}
In this section, we discuss the impact of some details of the model on the final performance. Some suggestions for selecting hyper-parameters based on relevant experimental results are also presented. The experiments follow the protocol of semi-supervised experiments, using 1\% labeled data and fine-tune evaluation.
\subsection{Clustering Methods}\label{Clustering Methods}
The clustering results are critical to the performance of ClusterCLHAR. Supposing that the clustering results are close to or equal to the true classification, the contrastive learning model will encode more significant differences in the samples of different classes to serve the downstream task better. Here we discuss the impact of using different clustering methods and corresponding clustering similarity measures on the final model performance when performing negative example masking. The clustering methods used in this subsection are K-means \cite{macqueen1967some} \cite{sculley2010web}, DBSCAN \cite{ester1996density}, Hierarchical clustering \cite{dasgupta2005performance} and BIRCH \cite{zhang1996birch}. The number of true classifications is used as the cluster number by default. The experiment results are shown in Table \ref{Tab03}.
\begin{table}
  \centering
  \caption{Clustering Methods}
  \label{Tab03}
  \setlength{\tabcolsep}{1.5mm}{
  \begin{tabular}[]{llccc}
    \toprule
      Method &Metric &USC-HAD &MotionSense &UCI-HAR\\
    \midrule
    K-means &euclidean &\textbf{78.09} &87.35 &86.03\\
    BIRCH &euclidean &68.41 &\textbf{87.97}  &\textbf{90.53}\\
    \multirow{2}{*}{Hierarchical.} &euclidean &76.31 &86.33 &89.19\\
    &cosine &76.41 &87.90 &89.61\\
    \multirow{2}{*}{DBSCAN} &euclidean &72.08 &86.21 &88.58\\
    &cosine &71.82 &85.95 &86.10\\
    
    \bottomrule
  \end{tabular}
  }
\end{table}
\par The experiment results show that the best performing clustering method on the USC-HAD dataset is K-means. It has a large gap with other methods, so k-means is chosen as the primary clustering method for this dataset. The best performing clustering method on the MotionSense dataset is Birch. But K-means was chosen as the primary clustering method for this dataset because it performs similarly to Birch and follows Occam's Razor. The best performing clustering method on the UCI-HAR dataset is BIRCH, and it has a larger gap with other methods, so it is chosen as the primary clustering method for this dataset. The difference in performance between the best and worst clustering methods was 9.68\%, 2.02\%, and 4.5\% for the three datasets, respectively. This result can demonstrate that the choice of clustering method has a crucial impact on the performance of ClusterCLHAR. Overall, we recommend using k-means as the first clustering method tried because it is simple, runs faster, and performs well under different distributions.

\subsection{Number of clusters}
 After determining the clustering method, the number of different clusters can also have a large impact on the model performance. We set 2 (batch size/512), 4 (batch size/256), 8 (batch size/128),16 (batch size/64), and 32 (batch size/32) as the number of clusters to explore their effect on the model performance. K-means was used as the clustering method for contrastive loss functions on the USC-HAD and MotionSense datasets, and BIRCH was used on the UCI-HAR dataset. The experiment results are shown in Table \ref{Tab04}.
\begin{table}
  \centering
  \caption{Number of clusters}
  \label{Tab04}
  \setlength{\tabcolsep}{1.5mm}{
  \begin{tabular}[]{cccc}
    \toprule
      Number of clusters &USC-HAD (12) &MotionSense (6) &UCI-HAR (6)\\
    \midrule
    True &\textbf{78.09} &87.35 &90.56\\
    2 &73.50 &85.92  &87.20\\
    4 &76.94 &86.98  &89.80\\
    8 &77.61 &87.11  &\textbf{90.73}\\
    16 &77.31 &87.19  &90.05\\
    32 &76.66 &\textbf{87.95}  &90.35\\
    \bottomrule
  \end{tabular}
  }
\end{table}
\par
The experimental results show that the difference between the best and worst cluster numbers on the three data sets is 4.59\%, 2.03\%, and 3.53\%, respectively. This result demonstrates that the choice of the number of clusters in the clustering method has a significant impact on the performance of the contrastive learning model. The choice of the number of clusters directly affects the definition of the number of negative pairs, thus also demonstrating that the definition of negative pairs is critical to model performance. In addition, we found that the number of clusters performed similarly when the true number of classifications was near, and the model performed worse when it was smaller. For this reason, we can conclude that when pre-training the unknown classification dataset for contrastive learning, we can choose the number of clusters that is larger than the estimated number of classifications. 

\subsection{Clustering Confidence}
In this subsection, we implement the method of Section \ref{Clustering Confidence} to control the proportion of normal samples in each clustering by setting the confidence level $\alpha$. The abnormal samples are set as a single class. We will discuss the performance on the downstream activity recognition task when different confidence levels $\alpha$ are chosen. It also verifies the difference between Cluster-NT-Xent ($\alpha=100$) and NT-Xent ($\alpha=0$). Here we set two cluster numbers, one using the correct classification number and the other at 16. The specific experiment results are shown in Table 3.
\begin{table}
  \centering
  \caption{Clustering Confidence}
  \label{Tab05}
  \setlength{\tabcolsep}{2mm}{
  \begin{tabular}[]{cccc}
    \toprule
      Confidence Level &USC-HAD &MotionSense &UCI-HAR\\
    \midrule
    cluster = true\\
    100 &\textbf{78.09} &87.35 &\textbf{90.56}\\
    95 &75.69 &\textbf{87.69}  &88.66\\
    90 &74.76 &86.32  &88.11\\
    80 &76.25 &86.67  &87.10\\
    0 &72.87 &85.49  &86.45\\
    \midrule
    cluster = 16\\
    100 &77.31 &87.19 &\textbf{90.05}\\
    95 &\textbf{77.48} &87.30  &89.77\\
    90 &76.56 &\textbf{87.48}  &89.86\\
    80 &76.71 &86.67  &89.27\\
    0 &72.87 &85.49  &86.45\\
    \bottomrule
  \end{tabular}
  }
\end{table}
\par
Overall, regardless of the number of clusters chosen, the model performs better with a large confidence level $\alpha$. In other words, the more the idea of Cluster-NT-Xent is followed, the better the performance is. It performs best on the MotionSense dataset with a confidence level of $\alpha$ of 95 but only improves by about 0.3\% compared to when $\alpha$ takes the value of 100. For the overall results, we speculate that this is due to the sensor data indicating easier clustering (higher confidence in the clustering results), for which adding a disconfidence operation to the clustering results (setting the proportion of anomalous samples) would have a negative effect on the model performance. Therefore, we recommend a preference for a confidence level of 100 when pre-training a new dataset.

\subsection{Batch size and epochs}
\begin{figure*}[htbp]
\centering
\subfigure[USC-HAD]{
\label{fig:subfig:e}
\begin{minipage}[t]{0.5\linewidth}
  \centering
  \includegraphics[width=3in]{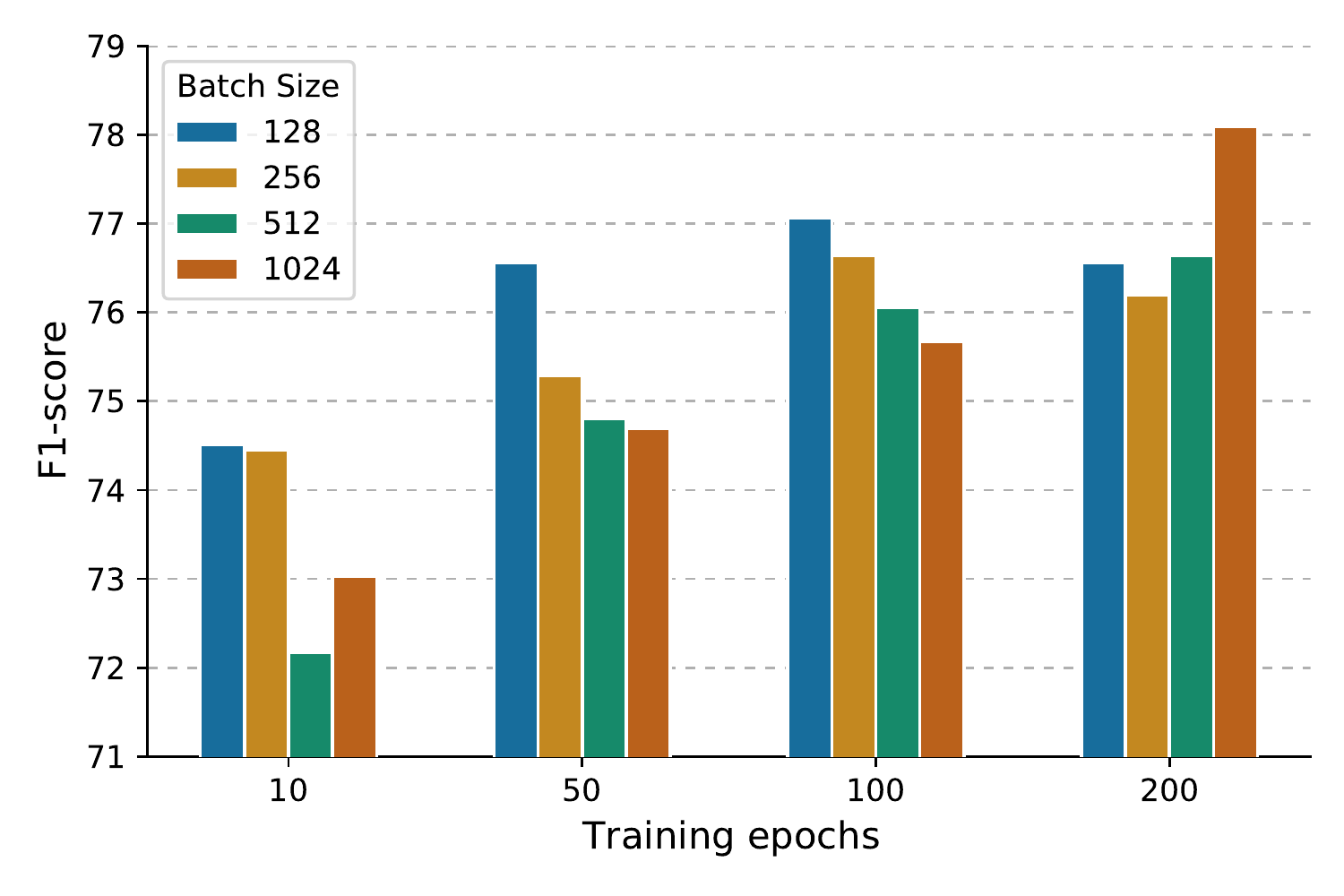}
\end{minipage}%
}%
\subfigure[MotionSense]{
\label{fig:subfig:f}
\begin{minipage}[t]{0.5\linewidth}
  \centering
  \includegraphics[width=3in]{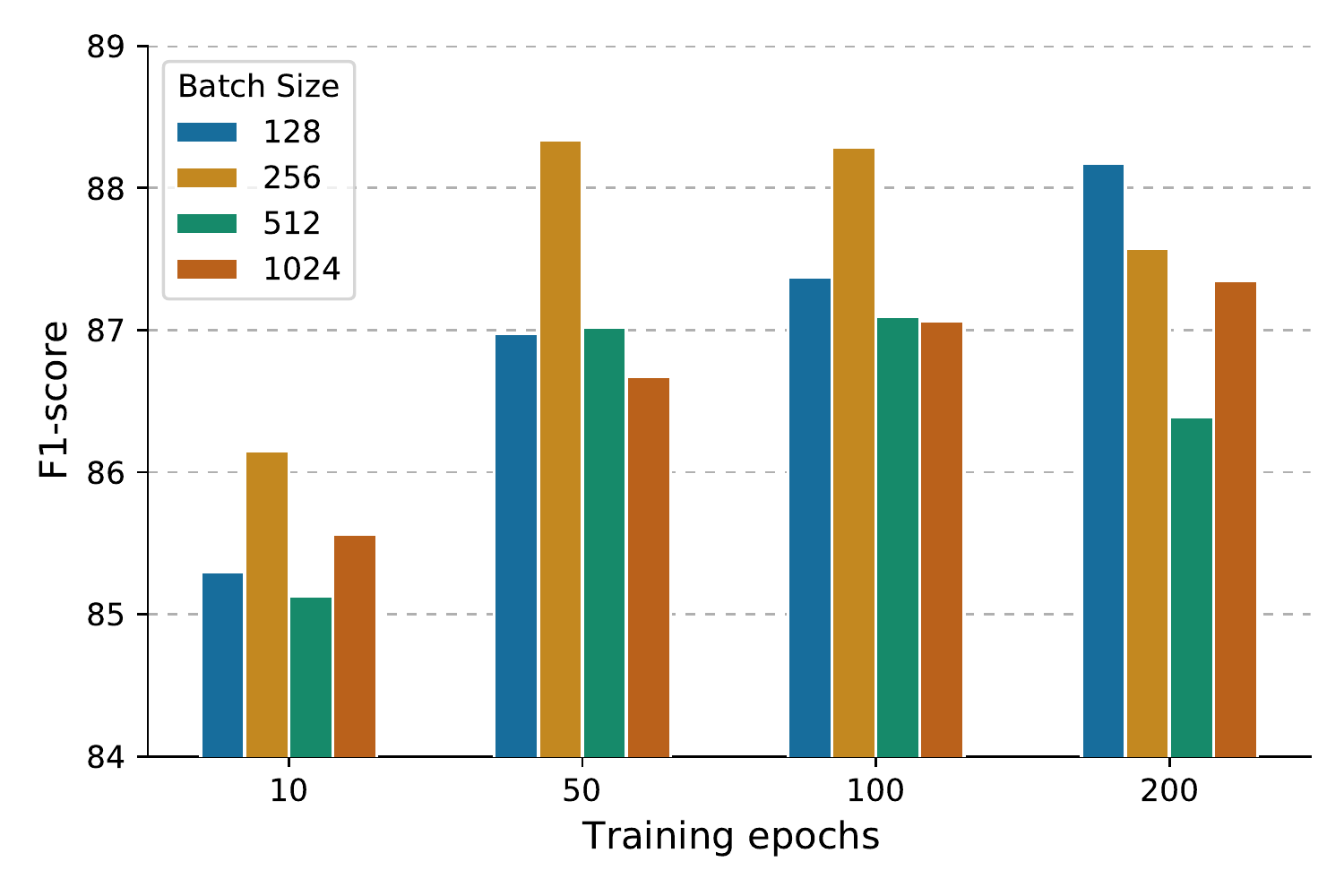}
\end{minipage}
}%
\caption{Batch size and epochs}
\label{fig:4}
\end{figure*}

This subsection will explore the impact of using different batch size and epochs in the pre-training phase of contrastive learning on model performance. The clustering method uses the conclusions of Section \ref{Clustering Methods}, and the number of clusters uses the true classification number. The experiment results are shown in Fig. \ref{fig:4}.
\par
The experiment results show that the two datasets have different sensitivities for different batch size and epochs, and the regularity is more difficult to find. With a determined batch size, the model may overfit as epochs increase. The selection of different batch sizes can also produce significant differences in model performance under deterministic epochs. As the batch size increases, the number of negative pairs also increases, which will increase the difficulty of clustering. We speculate that such an unstable regularity is that the batch size and epoch affect the performance of the clustering, which in turn affects the performance of the whole model. For this reason, we should carefully choose the batch size and training epoch when pre-training on other datasets.

\subsection{t-SNE}
\begin{figure*}[htbp]
\centering
\subfigure{
\rotatebox{90}{\scriptsize{~~~~~~~~~~~~~~~~~~~~\textbf{NT-Xent}}}
\begin{minipage}[t]{0.33\linewidth}
  \centering
  \includegraphics[width=1\linewidth]{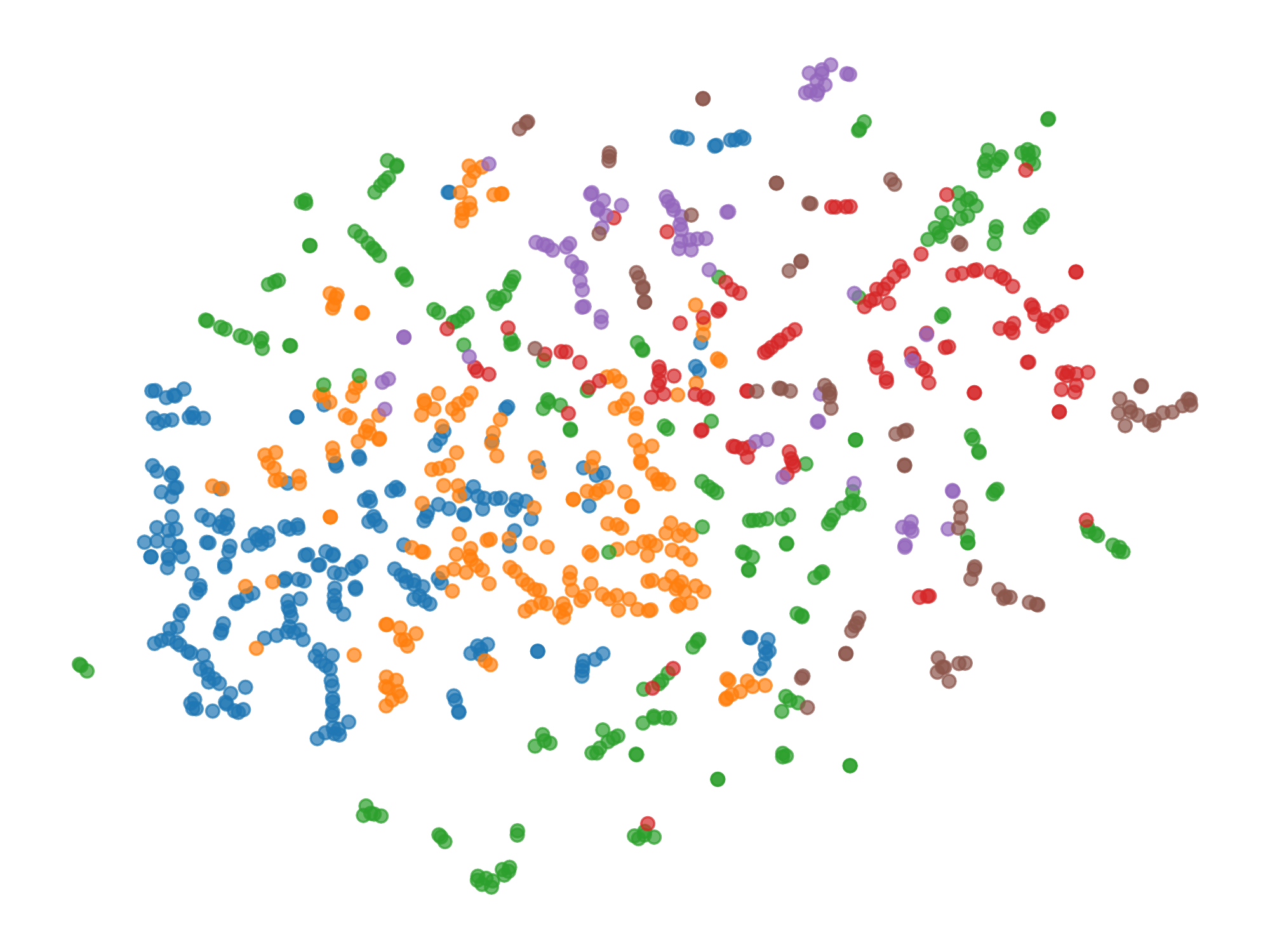}
\end{minipage}%
}%
\subfigure{
\begin{minipage}[t]{0.33\linewidth}
  \centering
  \includegraphics[width=1\linewidth]{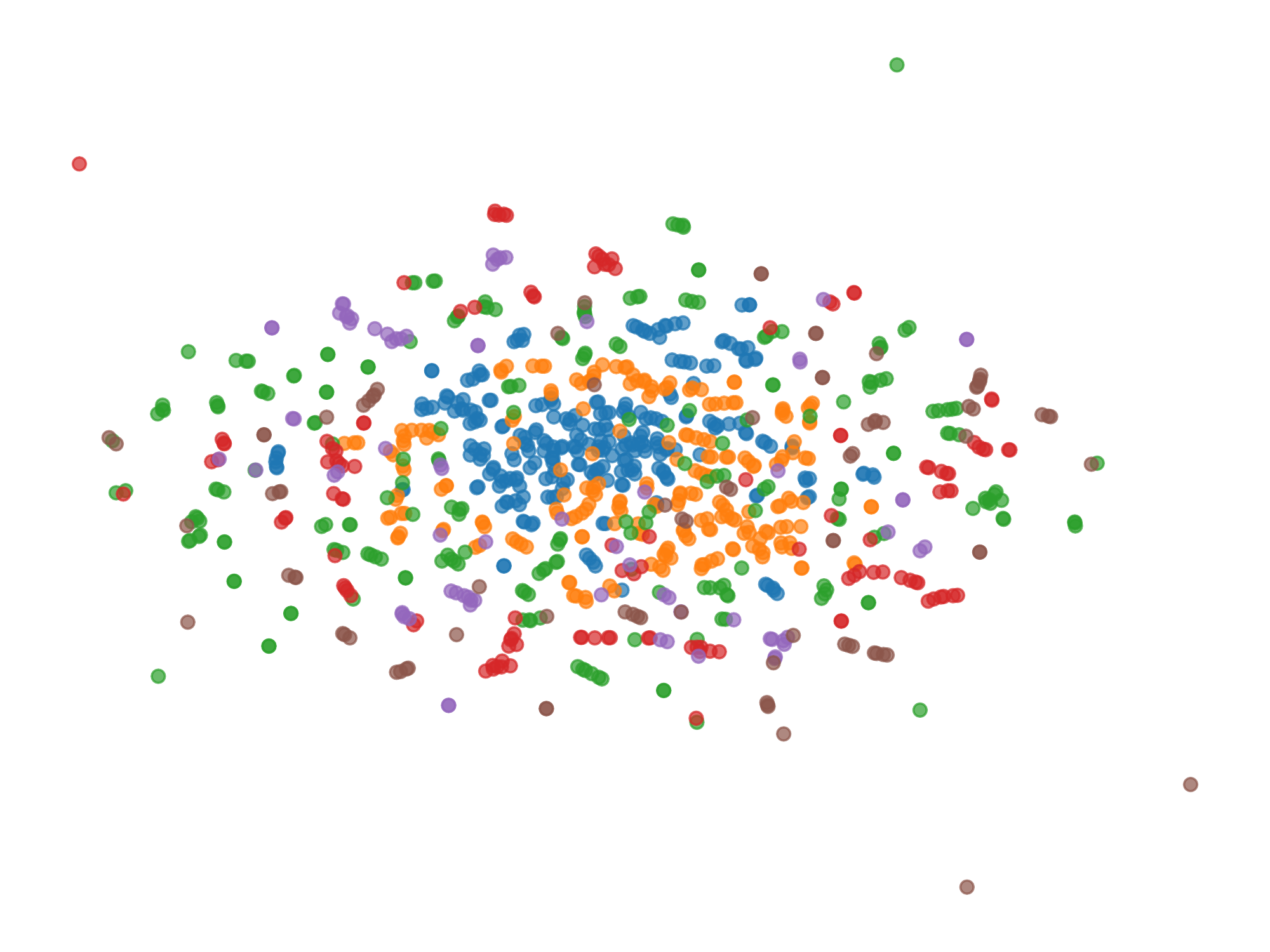}
\end{minipage}
}%
\subfigure{
\begin{minipage}[t]{0.33\linewidth}
  \centering
  \includegraphics[width=1\linewidth]{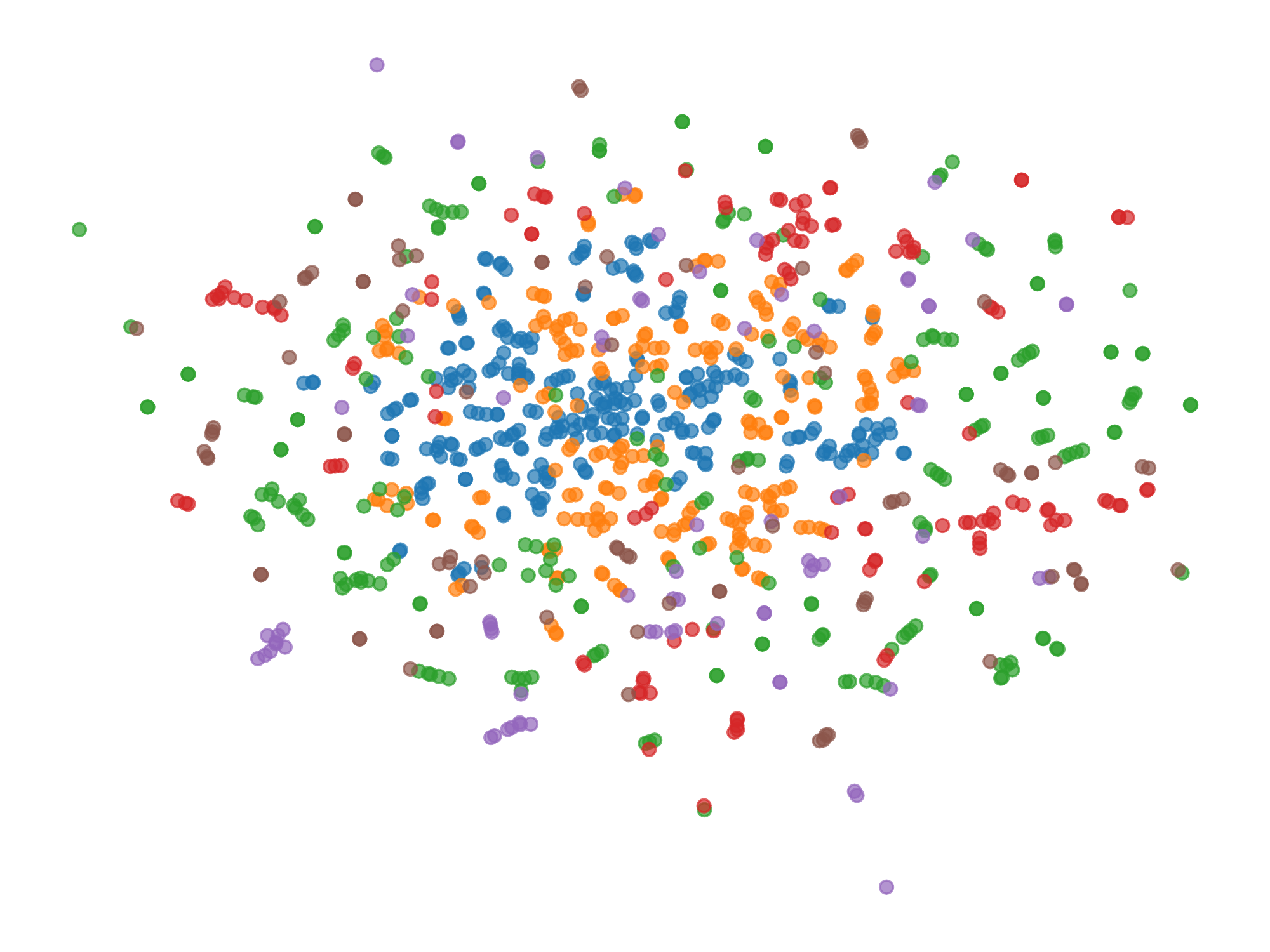}
\end{minipage}
}%

\vspace{-3mm}
\setcounter{subfigure}{0}

\subfigure[10 epochs]{
\rotatebox{90}{\scriptsize{~~~~~~~~~~~~\textbf{Cluster-NT-Xent}}}
\begin{minipage}[t]{0.33\linewidth}
  \centering
  \includegraphics[width=1\linewidth]{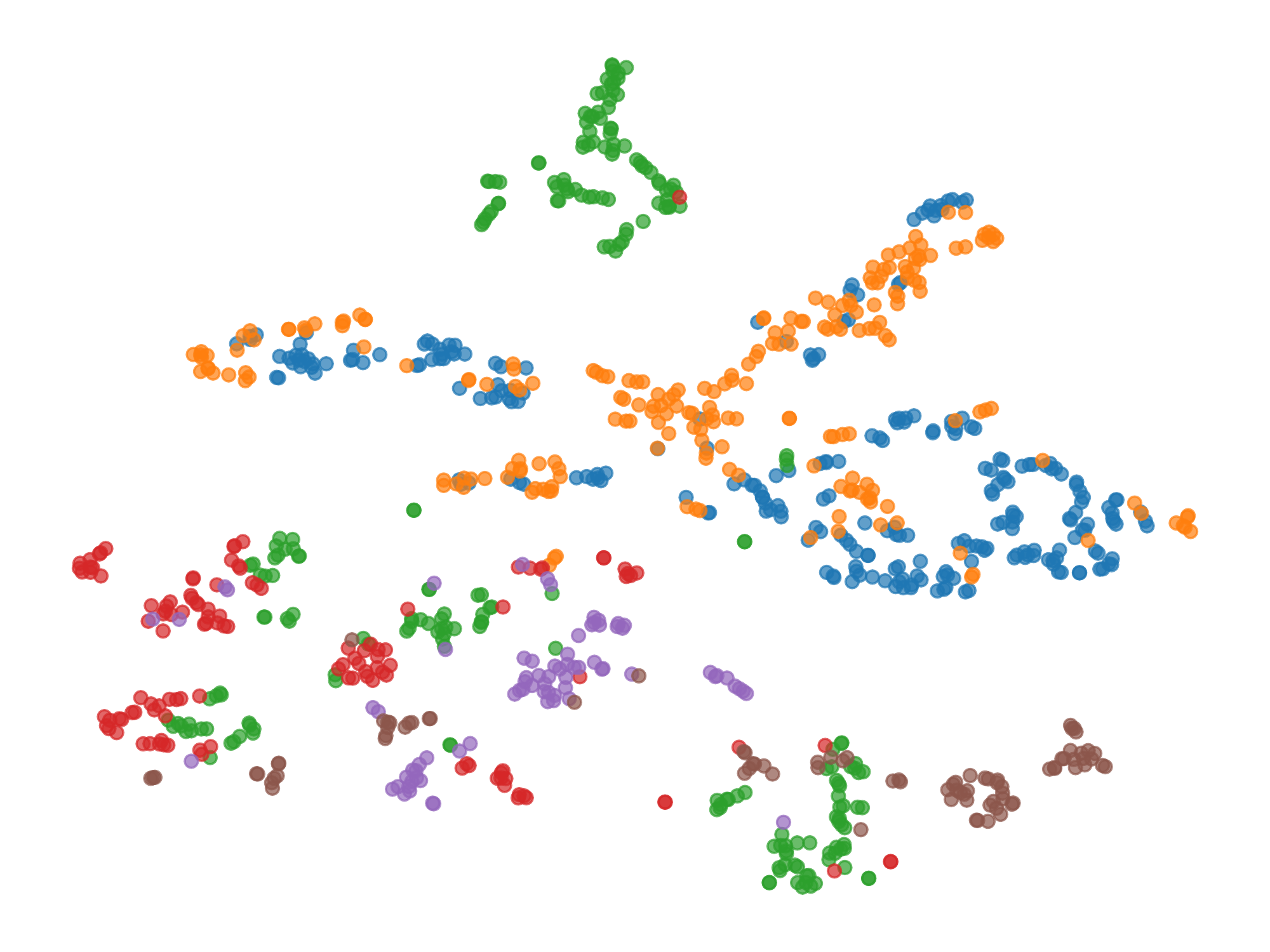}
\end{minipage}%
}%
\subfigure[100 epochs]{
\begin{minipage}[t]{0.33\linewidth}
  \centering
  \includegraphics[width=1\linewidth]{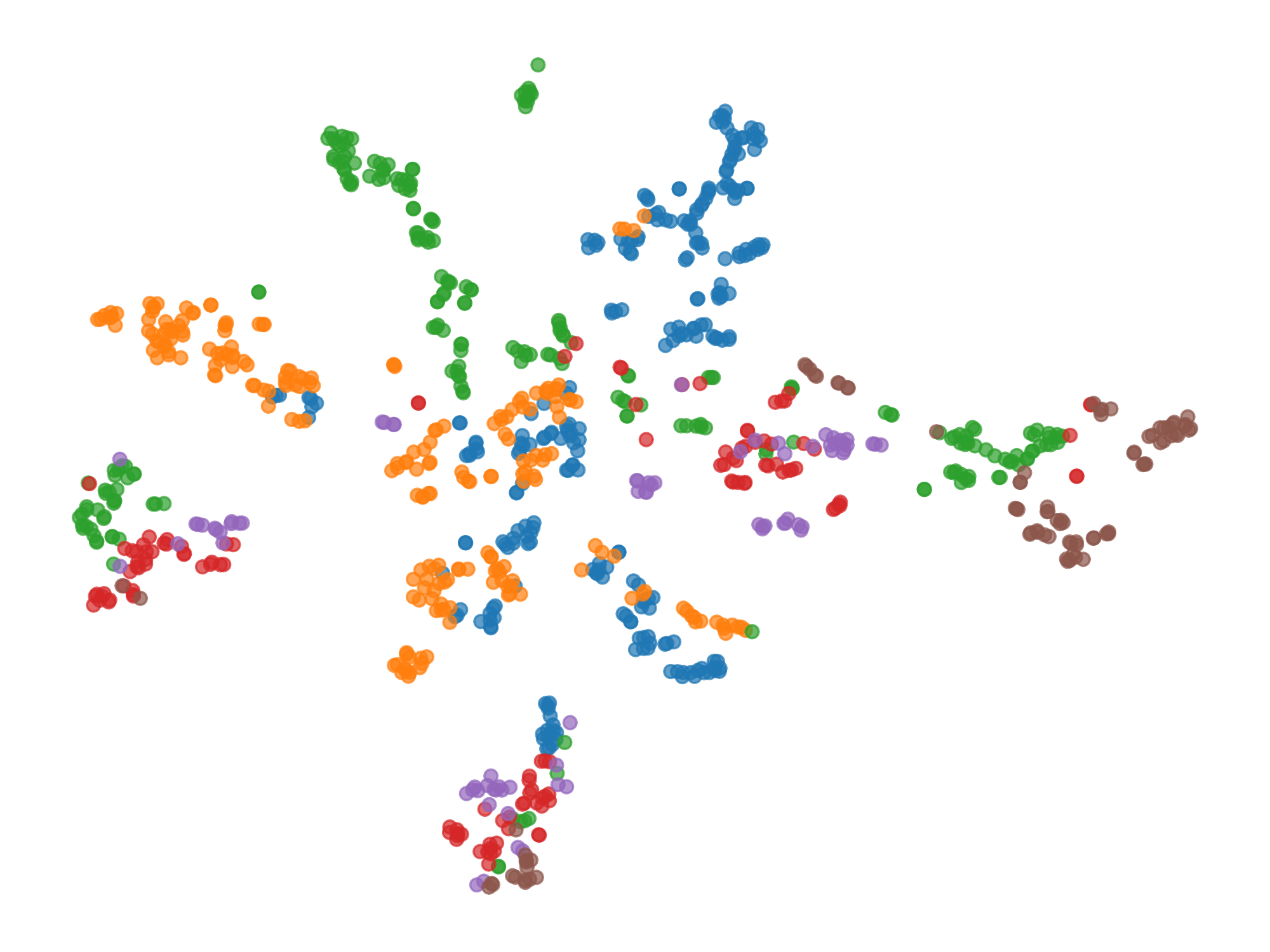}
\end{minipage}
}%
\subfigure[200 epochs]{
\begin{minipage}[t]{0.33\linewidth}
  \centering
  \includegraphics[width=1\linewidth]{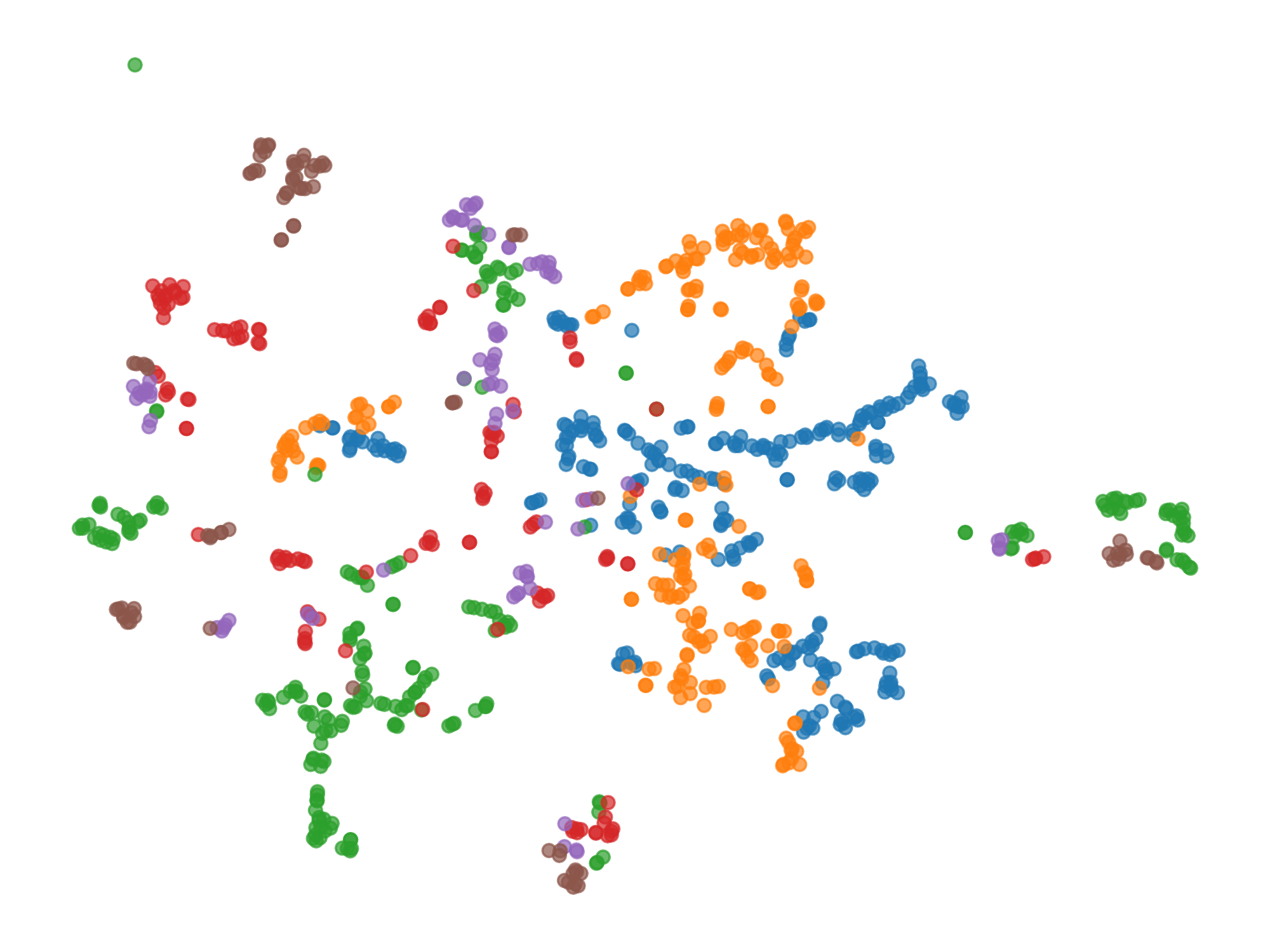}
\end{minipage}
}%
\caption{t-SNE}
\label{fig:5}
\end{figure*}

To show more graphically the effect of the pretext task in our proposed framework, we compared NT-Xent and Cluster-NT-Xent under t-SNE \cite{van2008visualizing} \cite{van2014accelerating} dimensionality reduction. On the MotionSense dataset, we randomly select 1000 samples of data to obtain a 96-dimensional representation (output of the projection header) by pre-training the model with contrastive learning. For these representations using the t-SNE method to reduce the dimensionality to 2 dimensions, the effect is obtained as shown in Fig. \ref{fig:5}.
\par
As can be seen from the figure, the NT-Xent distribution is rather chaotic, approximating that each sample is divided into a single class (similar to Fig. \ref{fig:subfig:a}). The above situation is in line with the idea of instance discriminations. Our proposed Cluster-NT-Xent somewhat aggregates the same class of representations together, which is very popular for downstream classification tasks. It can thus be demonstrated again that the introduction of clustering methods in the contrastive loss function can help the model present excellent performance in the downstream activity recognition task.

\section{Conclusion}
In this paper, we propose a contrastive learning framework that negative selection by clustering in HAR, called ClusterCLHAR. It outperforms all state-of-the-art work on self-supervised learning and semi-supervised learning for activity recognition tasks. The motivation of this paper is that the same class of samples will be considered as negative examples in the instance discrimination task, which will contradict the downstream classification task. The method in this paper is to introduce the clustering method acting on the sample representation, which is to not consider the same cluster samples as negative examples when calculating the contrastive loss. The paper concludes with a detailed discussion of the impact of some model details. And some suggestions on the choice of hyper-parameters are presented based on the relevant experiment results.
\par
There are also some shortcomings in this paper. Our proposed framework has limited improvement over supervised learning with a large amount of labeled data. In the discussion of the model details, we found that the performance of clustering can easily affect the final performance of the model. Based on the above issues, we decided to continue to optimize the definition of same-cluster samples in the contrastive learning framework in our future work. Real-life applications of the contrastive learning framework will also be investigated, such as making the most of the large amount of unlabeled data collected in real-time.

\section*{Acknowledgements}
This work is supported by the National Natural Science Foundation of China (61872038, 62006110).

\ifCLASSOPTIONcaptionsoff
\newpage
\fi

\bibliographystyle{IEEEtran}
\bibliography{main}

\begin{IEEEbiography}[{\includegraphics[width=1in,height=1.25in,clip,keepaspectratio]{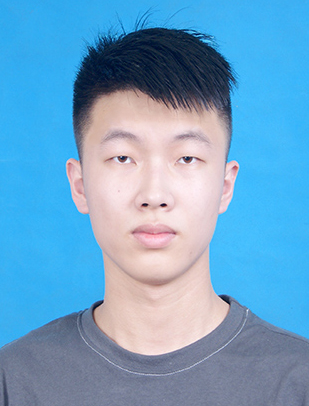}}]{Jinqiang Wang}
received his B.E. degree from Henan Normal University in 2020. He is currently a M.S. student in the School of Computer Science, University of South China. His research interests include intelligent perception and pattern recognition.
\end{IEEEbiography}

\begin{IEEEbiography}[{\includegraphics[width=1in,height=1.25in,clip,keepaspectratio]{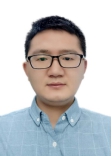}}]{Tao Zhu}
received the Ph.D. degree from University
of Science and Technology of China in 2015 and the BE
degree from Central South University in 2009. Then,
he worked as a post-Ph.D. and a lecturer in School of
Computer and Communication Engineering, University
of Science and Technology Beijing. Currently, he is
with University of South China. His research interests
include Evolutionary Computation and Internet of Things.
\end{IEEEbiography}

\begin{IEEEbiography}[{\includegraphics[width=1in,height=1.25in,clip,keepaspectratio]{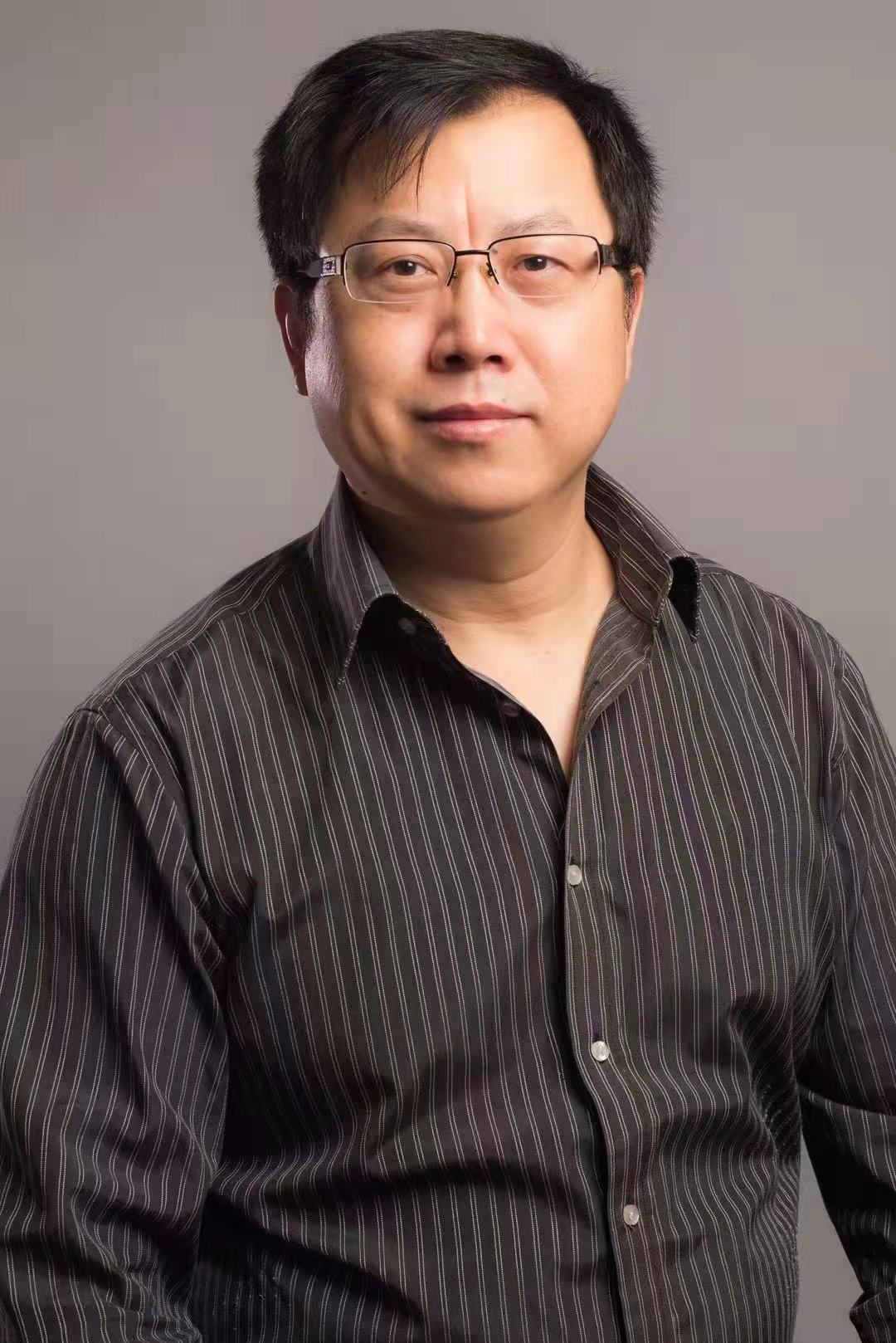}}]{Liming Luke Chen}
is Professor of Data Analytics in the School of Computing, Ulster University, UK. He received his BEng and MEng degrees at Beijing Institute of Technology, China, and DPhil on Computer Science at De Montfort University, UK. His current research interests include pervasive computing, data analytics, artificial intelligence, user-centred intelligent systems and their applications in smart healthcare and cyber security. He has published over 250 papers in the aforementioned areas. Liming is an IET Fellow and a Senior Member of IEEE.
\end{IEEEbiography}
\begin{IEEEbiography}[{\includegraphics[width=1in,height=1.25in,clip,keepaspectratio]{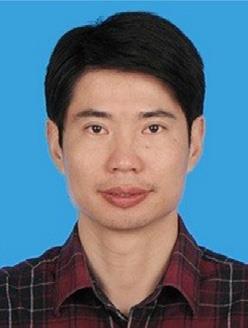}}]{Huansheng Ning}
received his B.S. degree from Anhui
University in 1996 and his Ph.D. degree from Beihang
University in 2001. He is currently a Professor and Vice
Dean with the School of Computer and Communication
Engineering, University of Science and Technology
Beijing and China and Beijing Engineering Research
Center for Cyberspace Data Analysis and Applications,
China, and the founder and principal at Cybermatics
and Cyberspace International Science and Technology
Cooperation Base. He has authored several books and
over 70 papers in journals and at international conferences/
workshops. He has been the Associate Editor of IEEE Systems Journal
and IEEE Internet of Things Journal, Chairman (2012) and Executive Chairman
(2013) of the program committee at the IEEE international Internet of Things
Conference, and the Co-Executive Chairman of the 2013 International Cyber
Technology Conference and the 2015 Smart World Congress. His awards include
the IEEE Computer Society Meritorious Service Award and the IEEE Computer
Society Golden Core Member Award. His current research interests include
Internet of Things, Cyber Physical Social Systems, electromagnetic sensing and
computing.
\end{IEEEbiography}

\begin{IEEEbiography}[{\includegraphics[width=1in,height=1.25in,clip,keepaspectratio]{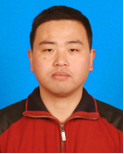}}]{Yaping Wan}
required Ph.D. degree from Huazhong University of Science and Technology. His research interests include big data causal inference.
\end{IEEEbiography}

\end{document}